\documentclass{article}
\usepackage{graphicx} 
\usepackage{multirow}
\usepackage{subcaption,amsmath}

\usepackage[preprint]{neurips_2026}

\usepackage[utf8]{inputenc} 
\usepackage[T1]{fontenc}    
\usepackage{hyperref}       
\usepackage{url}            
\usepackage{booktabs}       
\usepackage{amsfonts}       
\usepackage{nicefrac}       
\usepackage{microtype}      
\usepackage{xcolor}         

\title{Conditional generation of antibody sequences with classifier-guided germline-absorbing discrete diffusion}

\author{%
  Justin Sanders $^{1}$\thanks{Work performed during internship at Amazon Web Services} \\
  \texttt{jsander1@cs.washington.edu}
  \And
  Luca Giancardo $^{2}$ \\
  \texttt{lugianca@amazon.com}
  \And
  Lan Guo $^{2}$ \\
  \texttt{languo@amazon.com}
  \And
  Yue Zhao $^{2}$ \\
  \texttt{yuezhaom@amazon.com}  
  \And
  Kemal Sonmez $^{2}$ \\
  \texttt{ksonmez@amazon.com}
  \AND
  Nina Cheng $^{2}$ \\
  \texttt{cxinyun@amazon.com}  
  \And 
  Melih Yilmaz $^{2}$ \\
  \texttt{melihyz@amazon.com} 
\And
\\
$^1$ Paul G.\ Allen School of Computer Science and Engineering, University of Washington \\ \quad $^2$ Life Sciences, Amazon Web Services 
}

\begin{document}

\maketitle

\begin{abstract}
Antibody therapeutics are among the most successful modern medicines, yet computationally designing antibodies with desirable binding and developability properties remains challenging. While protein language models (pLMs) have emerged as powerful tools for antibody sequence design, existing approaches largely suffer from two key limitations: they predominantly memorize germline sequences rather than modeling biologically meaningful somatic variation, and they offer limited support for flexible classifier-guided conditional generation. We address these challenges through two primary contributions. First, we demonstrate that discrete diffusion fine-tuning achieves strong language modeling performance on antibody sequences while allowing for generation conditioned on any off-the-shelf classifier. Second, we introduce germline absorbing diffusion, a novel modification of the discrete diffusion noise process in which the germline sequence — rather than a masked sequence — serves as the absorbing state. This biologically motivated inductive bias restricts the model to learning the trajectory from germline to observed sequence, effectively excluding genetic variation and V(D)J recombination statistics from the learned distribution and dramatically mitigating germline bias. We show that germline diffusion improves non-germline residue prediction accuracy from 26\% to 46\%, approaching the theoretical upper bound set by true biological variability.  We then demonstrate the utility of our germline diffusion model on the conditional generation tasks of sampling antibodies with improved hydrophobicity and predicted binding affinity. On both tasks our model shows an improved tradeoff between class adherence and sample quality, significantly outperforming EvoProtGrad, a popular strategy to sample from pLMs with gradient-based discrete Markov Chain Monte Carlo. Notably, in an experiment to optimize the binding affinity of the clinical antibody Emibetuzumab to its antigen HGFR, our model generates designs with 48\% higher Boltz-2 predicted antibody-antigen complex scores on average, and produces the best predicted binder of any design. 
\end{abstract}

\section{Introduction}
Antibodies are a cornerstone of modern medicine, representing one of the most productive classes of therapeutic molecules \cite{mullard2021fda, lu2020development}. Their high binding specificity and compatibility with the human immune system make them ideal candidates for treating cancer, autoimmune diseases, and infectious pathogens. However, discovering and optimizing antibodies with the right combination of binding affinity, specificity, and developability properties remains a costly and time-consuming process, motivating the development of computational approaches to accelerate the design cycle.

Computational antibody design has historically relied on physics-based energy functions and structure-based methods \cite{adolf2018rosettaantibodydesign, sivasubramanian2009toward}, which, while powerful, are bottlenecked by the need for high-quality antigen structures and the prohibitive computational cost of exhaustive conformational search. The rise of protein language models (pLMs) has opened a complementary sequence-centric paradigm, where large-scale pretraining on hundreds of millions of protein sequences yields representations that implicitly encode rich structural and evolutionary information \cite{rives2021biological, lin2023evolutionary}. Antibody-specific pLMs such as AntiBERTy \cite{ruffolo2021deciphering}, AbLang \cite{olsen2022ablang}, and IgLM \cite{shuai2023generative} have demonstrated strong performance on tasks including liability prediction, humanization scoring, and complementarity-determining region (CDR) infilling. However, by training using standard unsupervised language modeling setups on large scale B-cell repertoires, these models often capture undesirable biases which limit their effectiveness. Natural B-cell sequences are shaped by many convolved biological processes: genetic variation in the V, D, and J gene segments, random rearrangement of these segments during development (V(D)J recombination), positive selection and affinity maturation to improve binding, and central/peripheral tolerance to avoid self-reactivity. While some of these processes are vital for a model to capture, others are extraneous to tasks of interest. 
Many antibody pLMs have been shown to primarily learn common genetic variants and V(D)J recombination patterns \cite{olsen2024ablang2, matsen2025separating}.
As a result, much of their apparent performance improvement over generic pLMs derives from memorizing germlines rather than accurately modeling the biologically meaningful variation that distinguishes functional antibodies — a critical limitation for \emph{de novo} design and directed evolution campaigns. 

Beyond the germline bias problem, existing antibody generative models, especially those allowing for conditional generation guided by another model (e.g. classifier-guided generation with a developability or binding affinity predictor), predominantly operate under autoregressive, single-step masked prediction, or gradient-guided Markov Chain Monte Carlo (MCMC) frameworks \cite{madani2023large, shuai2023generative, emami2023plug}. While autoregressive models carry inductive biases appropriate for natural language, they may be less well suited for antibodies since biological sequences do not necessarily carry the same sequential dependency on context. 
Additionally, and more importantly, these approaches can only be guided by classifiers which are fully differentiable with respect to the input sequence, severely limiting the types of models that can be used for guidance. Furthermore, even when they are computable, these gradients may not be semantically meaningful when working in a discrete domain. 
\vspace{-4pt}

\begin{figure*}[!htbp]
  \centering
    \includegraphics[height=3.0in]{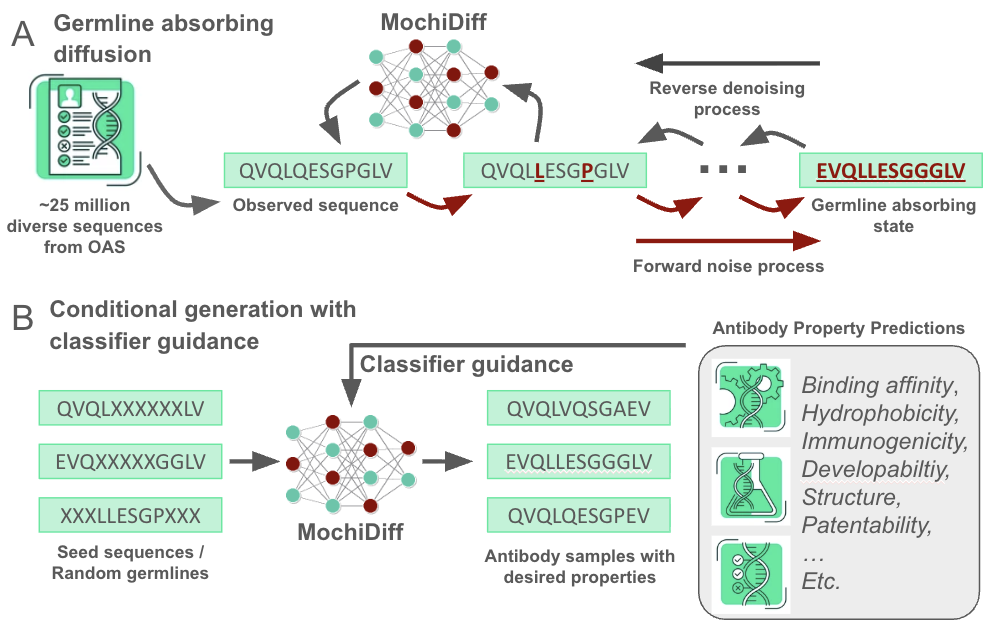} 
  \caption{{\bf Discrete diffusion antibody protein language model with germline absorbing state}
    (A) Our model is trained to denoise germline sequences into observed antibody sequences using the score entropy discrete diffusion (SEDD) framework. (B) Once trained our model allows for \textit{de novo} generation and directed evolution of antibody sequences conditioned on arbitrary classifiers.   
  }
  \label{fig:abstract}
  \vspace{-2pt}
\end{figure*}

Discrete diffusion models offer a compelling alternative \cite{austin2021structured, lou2024discrete}. By framing generation as the gradual reversal of a noise process over discrete sequence tokens, they naturally support flexible conditioning, efficient infilling, and straightforward classifier-guided generation. 
The inductive bias of forward diffusion is well-suited to biological sequences shaped by natural selection. Discrete diffusion models treat sequences as the result of a generative process where sequences undergo random substitutions with a bias towards regions of high likelihood. This is analogous to individual mutations gradually increasing fitness through natural selection. 

In this work, we present a novel discrete diffusion antibody pLM to address these opportunities. Our model is built on the Score Entropy Discrete Diffusion (SEDD) framework \cite{lou2024discrete} and leverages the ESM-2 transformer architecture \cite{lin2023evolutionary}, initialized from pretrained weights on general protein sequences and fine-tuned with a diffusion objective on over 25 million diverse B-cell receptor sequences from the Observed Antibody Space (OAS) database \cite{kovaltsuk2018observed, olsen2022oas}. To mitigate the germline bias problem, we employ a novel framework for discrete diffusion where the germline sequence is used as the absorbing state of the noise process. By explicitly pre-conditioning on the germline in this way, we hypothesize that the model will devote more capacity to learning aspects of the distribution of B-cell receptor sequences which are the most relevant for therapeutic design. Additionally, by more faithfully representing the true generative process underlying the data, we introduce a favorable inductive bias to the model. Evaluating our model on language modeling metrics, we find that diffusion models are well suited as antibody pLMs, and our germline absorbing diffusion setup significantly reduces germline bias. In downstream  
\textit{in silico} conditional generation tasks, our new model, which we call MochiDiff, successfully designs sequences with desirable developability and binding properties. 

\section{Background and related work}

\paragraph{Antibody design}
Antibodies are proteins whose target specificity is primarily determined by six CDRs, of which CDR-H3 is the most diverse and functionally critical \cite{chothia1989conformations, north2011new}. Antibodies can be represented with \textit{sequences} of letters that describe amino acids and/or a list of 3D atom positions that describe their physical \textit{structure}. Traditional computational design relied on physics-based energy functions in structure space \cite{adolf2018rosettaantibodydesign, sivasubramanian2009toward}, which are computationally expensive and require high-quality antigen structures. Deep learning has substantially relaxed these constraints with recent diffusion-based methods such as DiffAb \cite{luo2022antigen}, AbDiffuser \cite{martinkus2023abdiffuser}, RFdiffusion \cite{watson2023novo}, and Chroma \cite{ingraham2023illuminating} 
enabling both unconditional and conditionally guided backbone generation. 
Despite these advances, limited availability of resolved antigen structures, and inherent computational costs due to the higher-dimensional space of atom positions still bottleneck structure-based design, motivating pLM-based approaches that can operate from sequence data alone.

\paragraph{Protein language models and conditional generation}
Large-scale pretraining on protein sequence databases has produced pLMs that encode rich evolutionary, structural, and functional information in learned sequence representations \cite{rives2021biological, 
elnaggar2021prottrans}. ESM-2 \cite{lin2023evolutionary}, trained on hundreds of millions of sequences with a masked language modeling objective, produces residue-level embeddings from which secondary structure, solvent accessibility, and contact maps are recoverable via linear probing. This has made pLMs attractive not just as encoders but as generative priors for conditional sequence design, with two main paradigms emerging.

The first paradigm fine-tunes pLMs on domain-specific antibody data to capture the immunoglobulin-specific statistics that distinguish antibodies from generic proteins. AntiBERTy \cite{ruffolo2021deciphering} and AbLang \cite{olsen2022ablang} are BERT-style \cite{devlin2019bert} masked language models pretrained exclusively on antibody repertoire sequences, demonstrating superior performance on liability prediction, humanization scoring, and CDR infilling compared to general pLMs. 
However, AbLang-2 \cite{olsen2024ablang2} subsequently showed that much of this improvement derives from memorizing common germline sequences and V(D)J recombination patterns rather than accurately modeling the biologically meaningful somatic variation that distinguishes functional antibodies, a critical limitation for conditional generation of novel therapeutic sequences. Some of these antibody pLMs support conditional generation by prepending discrete property or annotation tokens during training. ProGen \cite{madani2023large} and ProGen2 \cite{nijkamp2022progen2} condition an autoregressive model on functional family annotations to enable controllable generation across broad protein families. Within the antibody domain, IgLM \cite{shuai2023generative} adopts the same causal language modeling framework conditioned on chain type and species tokens, enabling targeted antibody sequence generation. While effective, architectural conditioning requires that the desired property be pre-specified at training time and accompanied by sufficient labeled training data, conditions that rarely hold for many biophysical properties central to therapeutic design. Crucially, because conditioning is locked in at training time, none of these models natively support flexible classifier-guided generation. 

The second paradigm uses the pLM's learned likelihood landscape to guide generation post-hoc, without modifying the model itself. EvoProtGrad \cite{emami2023plug} provides a composable gradient-based MCMC framework that combines one or more fitness predictors with pLM pseudo-log-likelihood scores to guide discrete sequence optimization. This allows any differentiable classifier to be plugged in without retraining the pLM. However, gradient-based MCMC over discrete sequences requires continuous relaxations that introduce  error, and the requirement for differentiable guidance classifiers severely limits what models can be used out-of-the-box.

\paragraph{Discrete diffusion}
Diffusion probabilistic models \cite{sohl2015deep, ho2020denoising, song2020score} have achieved remarkable success in continuous domains such as image \cite{rombach2022high}, audio \cite{kong2020diffwave}, and protein structure generation \cite{jumper2021highly, yim2023se}; their extension to discrete sequences requires fundamentally different forward-process formulations. Austin et al. introduced D3PM \cite{austin2021structured}, a unifying framework for discrete diffusion that generalizes multinomial corruption processes. Concurrent work on multinomial diffusion \cite{hoogeboom2021argmax} demonstrated categorical diffusion for text and image segmentation. A key finding across these works is that the absorbing diffusion process—wherein tokens are progressively replaced by a special \texttt{[MASK]} token—tends to be particularly effective as it serves as a generalization of single-noise level BERT-style masked language modeling to multiple continuous noise levels \cite{he2022masked}. 
SEDD \cite{lou2024discrete} extends the score matching framework to discrete state spaces by parameterizing score ratios of the data distribution rather than continuous score functions. SEDD admits an analytically tractable score entropy training objective and achieves GPT-2-competitive perplexity on language benchmarks. Additionally, its simple formulation supports both efficient and flexible infilling without any task-specific fine-tuning, a property especially well-suited to CDR loop co-design, and straightforward classifier-guided generation \cite{schiff2025simple}, a property invaluable for \textit{de novo} and directed evolution antibody design campaigns. In the protein domain, EvoDiff \cite{alamdari2023protein} demonstrated that discrete diffusion over amino acid vocabularies with ESM-2 embeddings can generate evolutionarily plausible and structurally diverse proteins in sequence space alone, while DpLM \cite{wang2024dpLM} showed that pretraining a discrete diffusion model on large protein databases yields a flexible generative prior amenable to conditional fine-tuning. Our work adopts the SEDD framework as its generative backbone, adapting it to the antibody setting with methodological modifications informed by B-cell developmental biology.

\section{Methods}
\paragraph{Germline discrete diffusion for protein sequences}
Our model's discrete diffusion adapts the SEDD approach described by Lou et al. \cite{lou2024discrete} by using a novel form of  absorbing state that exploits our understanding of the generative process underlying observed biological sequences. We consider the discrete diffusion process:
$$p(y_{t + \Delta t} | x_t) = Q_t(y,x)\Delta t$$
where  $x_t$ is the current state, $y_{t+\Delta t}$ is a candidate next state, $p(x_0) = p_{data}$, and $Q$ is the diffusion matrix. This process gradually transforms our data distribution at $p_0$ to the stationary distribution of the transition matrix $Q$.
The time reversal of this process has diffusion matrix $\bar{Q}$ given by: 
$$
\bar{Q_t}(y,x) =  \frac{p_t(y)}{p_t(x)}Q_t(x,y) \qquad \bar{Q_t}(x,x) = - \sum_{y \neq x}\bar{Q_t}(y,x)
$$
This describes a process which gradually maps the stationary distribution of $Q$, $p_T(x)$ where $T$ is larger than the mixing time of $Q$, back to the data distribution $p_0(x)$. In theory, the transition matrix $Q$ of the diffusion process is of size $n^d \times n^d$, where $n$ is the vocabulary size and $d$ is the sequence length. However, if we consider each token to be perturbed independently, it factors to tractable $n \times n$ token-level matrices. In nature, variants occur independently, making this assumption reasonable. 

In practice, we test two easy to compute matrices $Q$, yielding either a uniform stationary distribution or one with a predefined absorbing state. For biological sequences, this is not only a reasonable assumption but a desirable inductive bias - biological sequences undergoing natural selection accumulate mutations one residue at a time. Typically, discrete diffusion models are trained with one of two easy to compute matrices $Q$, yielding either a uniform stationary distribution or one with an all [MASK] token absorbing state. Here, we propose a novel form of the latter that exploits our understanding of the generative process underlying observed biological sequences. Our goal is to start generation from a germline sequence, preventing the model from learning germline statistics and focusing it on post-recombination variation. We do this by modifying the absorbing diffusion process to have the germline sequence (as annotated in OAS) as the absorbing state, instead of the typical all [MASK] token absorbing state, so the model learns only the trajectory from germline to observed sequence, and not the distribution of germlines.

To enable sampling from $p_{data}$ using the reverse process above, we need the ratio $\frac{p_t(y)}{p_t(x)}$, dubbed the concrete score. We train a neural network $S_\theta(x,t)$, to approximate the concrete score using training examples $x_t$, generated by running the forward diffusion process on training data $x_0$, and optimizing for the denoising score entropy loss function introduced by Lou et al. \cite{lou2024discrete}:
\[
    \mathcal{L}_{dse}(x_t, S_\theta) =  \sum_{i=1}^{d}\sum_{y=1, y \neq x_t^i}^{n} Q_t(x_t,y)(S_\theta(x_t,t)_y^i - \frac{p(y^i|x_0)}{p(x_t^i|x_0)}log(S_\theta(x_t,t)^i_y))
\]
where $d$ is the length of the sequence $x$ and $n$ is the number of tokens in the models alphabet. The ratio $\frac{p(y^i|x_0)}{p(x_t^i|x_0)}$ can be computed in closed form for appropriate choices of the diffusion matrix $Q$. For our germline absorbing diffusion process, this entails a modification to the standard absorbing diffusion:  
$$p(x_t^i|x_0) = e^{\sigma(t)Q}x_0^i = e^{-\sigma(t)}x_0^i + (1 - e^{-\sigma(t)})x_{Germ}^i$$
Where $\sigma(t)$ is the total noise level at time $t$ and $x_{Germ}$ is the germline sequence corresponding to $x$. To simplify training our absorbing diffusion models, we use the reparameterized absorbing discrete diffusion (RADD) trick presented by Ou et al \cite{ou2024absorbing} to give our model only analytical time dependence, removing the need to parameterize our network with the timestep $t$. 

To implement SEDD, we use a transformer architecture for $S_\theta(x,y,t)$, where for a given sequence $x$ and timepoint $t$ the model outputs a $d \times n$ matrix of likelihood ratios describing the score ratio between $x$ and each $y$ within edit distance 1 of $x$. This allows us to directly use the ESM-2 transformer model architecture along with its pretrained weights learned from general protein sequences. 

\paragraph{Dataset and model training}
To train our model, we use the observed antibody space (OAS) dataset of over 1.1 billion B-cell receptor sequences, which is ubiquitous in the field \cite{kovaltsuk2018observed, olsen2022oas}. To retain additional sequence metadata, including the IgBLAST-assigned germline sequences, we reprocess the dataset ourselves by filtering to require complete V(D)J annotations, no early stop codon or frameshift, productive sequences only, no unassigned ‘X’ residues in the sequence, and no missing conserved Cysteines. After applying these filtering rules there are 337 million total sequences remaining. We then apply a final clustering step on the filtered data using mmseqs linclust with a strict 70\% sequence identity cutoff \cite{steinegger2017mmseqs2, steinegger2018clustering}, ensuring high sequence diversity. This gives a final dataset of 25.6M sequences. From this, validation and test datasets of 800,000 sequences each are held out, ensuring at most an 80\% sequence similarity between germlines. 

Our core model is based on the ESM-2 architecture, with a slight modification to allow for conditioning on the diffusion timestep $t$ when required for uniform diffusion.  To minimize architectural changes, we use in-context conditioning with a sinusoidal time embedding with learned frequencies, rather than the AdaLN approach widely used in diffusion transformers \cite{peebles2023dit}. By using the same architecture as ESM-2, we can initialize our model with ESM-2 weights pre-trained with large scale generic protein data. For all experiments we use the time-conditioned ESM-2 650M parameter encoder initialized with the pretrained weights.

\paragraph{Unconditional sampling from the trained model}
Having trained a model $S_\theta$, we can perform unconditional sampling from $p_{data}$ by sampling from $p_t$ and simulating the reverse diffusion process using the learned concrete score. We use the Tweedie decoding strategy to sample from the reverse process, which has been shown to be optimal \cite{lou2024discrete}:
$$p(y|x_t) = (e^{(\sigma(t-\Delta t) - \sigma (t))Q}S_\theta (x_t,t)_y) e^{(\sigma(t) - \sigma (t-\Delta t))Q}(x_t,y)$$
where $\Delta t$ is the discrete time step size. There is an inherent tradeoff between sample efficiency and quality, with smaller $\Delta t$ yielding higher quality but more expensive samples.

\paragraph{Conditional sampling with classifier guidance}
A key advantage of the SEDD formulation is the ease with which sampling can be guided by a classifier to satisfy properties of interest. 
To sample from a conditional distribution $p(x|c)$ for some condition of interest $c$, we can use the predictions of a pre-trained classifier which predicts $f_\Omega(x) = p_t(c|x)$ by decomposing the concrete score during sampling into conditional and unconditional components as follows:
    \begin{equation}
  \frac{p_t(x_1|c)}{p_t(x_2|c)} = \frac{p_t(x_1)}{p_t(x_2)} \frac{p_t(c|x_1)}{p_t(c|x_2)} = S_\theta (x_2,t)_{x_1} \frac{f_\Omega(x_1)}{f_\Omega(x_2)} 
  \label{eq:auroc}
\end{equation}
Notably, this setup only requires the logits of the guidance classifier, as opposed to gradients. This is an advantage for discrete spaces such as proteins, where local gradients may not be meaningful. It also makes the approach completely agnostic to the architecture of $f_\Omega$, unlike other methods such as EvoProtGrad, which requires computable gradients with respect to the input. However, this comes with the downside of requiring $L * V$ forward passes of the classifier at each timestep during sampling, where $L$ is the length of the sequence and $V$ is the vocab size. This constrains guidance to relatively small models. Additionally, in practice the classifier logits typically have to be re-scaled by an empirical guidance strength parameter to appropriately balance the two terms. 

Another difficulty with applying classifier guidance is the requirement to parameterize the classifier with the timestep $t$. For the uniform and [MASK] absorbing models, this necessitates training a new classifier from scratch on noised data. Otherwise, noised samples will quickly fall well out of distribution. Another key advantage of our germline absorbing diffusion model is that even the fully 'noised' sequences are real antibodies, and thus in-distribution for a pre-trained classifier. This allows arbitrary property prediction models to be used for guidance off the shelf. 

\begin{table}[!htbp]
\centering
\begin{tabular}{ccc}
\hline
\textbf{Model}                                                                     & \begin{tabular}[c]{@{}c@{}}Perplexity $\downarrow$\\ \end{tabular} & \begin{tabular}[c]{@{}c@{}}Non-germline \\ accuracy $\uparrow$ \end{tabular} \\ \hline
ESM-2 \cite{lin2023evolutionary}                                                                               & 3.509                                                                     & 0.193                                                            \\
\begin{tabular}[c]{@{}c@{}}ESM-2 finetuned (MLM)\end{tabular}                    & 1.448                                                                     & 0.247  \\
AbLang-2 \cite{olsen2024ablang2}                                                                          & 1.875                                                                     & 0.229                                                            \\
IgLM \cite{shuai2023generative} & 1.411                                                                     & 0.257      
\\ \hline
\begin{tabular}[c]{@{}c@{}}MochiDiff uniform (ours)\end{tabular}           & \textbf{\textless{}= 1.351}                                               & \textbf{0.261}                                                            \\
\begin{tabular}[c]{@{}c@{}}MochiDiff absorbing (ours)\end{tabular}         & \textless{}= 1.467                                                        & 0.255                                                            \\
\begin{tabular}[c]{@{}c@{}}MochiDiff germline (ours)\end{tabular}          & \textbf{\textless{}= 1.293*}                                              & \textbf{0.463*}                                                   \\ \hline
\begin{tabular}[c]{@{}c@{}}Nearest neighbor baseline \end{tabular}    & NA                                                                        & 0.127                                                            \\
\begin{tabular}[c]{@{}c@{}}Nearest neighbor upper-bound \\ \end{tabular} & NA                                                                        & 0.538*                                                
\end{tabular}
  \caption{{\bf Benchmarking our diffusion model against other protein language models on antibody sequences.}
    Comparison of our models to other antibody pLMs on  perplexity and ability to accurately predict non-germline residues for a test set of held-out germlines. For diffusion models, an evidence-based upper bound on perplexity is reported. The best performing method(s) in each column are indicated in bold. Methods which take in additional input information are indicated with a star, indicating that perplexities are not directly comparable. Arrows indicate whether lower values ($\downarrow$) or higher values ($\uparrow$) are better for each metric.
    \vspace{-19pt}
  }
  \label{table:bench}
\end{table}

\section{Results}
\vspace{-2pt}
\paragraph{MochiDiff achieves strong language modeling performance }
We first evaluate our diffusion antibody pLMs using standard language modeling metrics. We consider three different variants of our MochiDiff model: a uniform diffusion model, a MASK absorbing state model, and a germline absorbing state model. We compare these to four existing approaches: an off-the-shelf ESM-2 650M model, ESM-2 finetuned on the same antibody specific training set but using the standard masked language modeling (MLM) loss used during pretraining, IgLM, an autoregressive antibody-specific pLM, and AbLang-2, another antibody pLM trained on OAS which uses a focal loss to mitigate the germline bias. 
Because AbLang-2 and IgLM use different train-test splits, there is a chance of some train sequence overlap artificially inflating their performance. 

For each sequence, we first estimate the test set perplexity. For the masked language models, we use  pseudo-perplexity, calculated by masking each token in the sequence individually. For our diffusion models, the perplexity cannot be calculated exactly, but we can derive an evidence based upper-bound which is expected to be tight in practice \cite{campbell2022continuous}. As expected, we find that finetuning on antibody specific data significantly lowers perplexity from 3.509 to 1.448 for ESM-2 (Table \ref{table:bench}).
Evaluating our diffusion models, we find that uniform diffusion outperforms MLM as a finetuning task, achieving a test set perplexity $\leq$ 1.351, also outperforming AbLang-2 (1.875) and IgLM (1.411). 
The MASK absorbing state diffusion model obtains perplexity $\leq$ 1.467, comparable to the MLM model.
Finally, our germline absorbing state model achieves by far the lowest perplexity of $\leq$ 1.293, however this perplexity is conditional since the model receives the germline sequence as additional input.

\begin{table}[!htbp]
\centering
\resizebox{\textwidth}{!}{%
\setlength{\tabcolsep}{3pt}
\renewcommand{\arraystretch}{1.3}
\footnotesize

\begin{tabular}{cccccccccccccc}
\textbf{} 
 &
   &
  \multicolumn{3}{c|}{V gene class} &
  \multicolumn{3}{c|}{Hydrophobicity} &
  \multicolumn{6}{c}{HGFR Binding} \\ \cline{3-14}
Model &
  \begin{tabular}[c]{@{}c@{}}Guid.\\ str.\end{tabular} &
  \begin{tabular}[c]{@{}c@{}}Class\\ adh. $\uparrow$ \end{tabular} &
  \begin{tabular}[c]{@{}c@{}}NN\end{tabular} &
  \multicolumn{1}{c|}{Div.} &
  \begin{tabular}[c]{@{}c@{}}kcal/\\ mol $\downarrow$ \end{tabular} &
  \begin{tabular}[c]{@{}c@{}}NN\end{tabular} &
  \multicolumn{1}{c|}{Div.} &
  \begin{tabular}[c]{@{}c@{}}Clf. $\uparrow$\\ p(bind)  \end{tabular} &
  \begin{tabular}[c]{@{}c@{}}Boltz-2 $\uparrow$\\ p(bind)  \end{tabular} &
  ipTM $\uparrow$ &
  pLDDT $\uparrow$ &
  Div. &
  \begin{tabular}[c]{@{}c@{}}Dist.\\ seed\end{tabular} \\ \hline
\multirow{4}{*}{\begin{tabular}[c]{@{}c@{}}MochiDiff\\ uniform\\ (ours)\end{tabular}} &
  None  & 0.09 & 0.94 & \multicolumn{1}{c|}{41.7} & 0.070 & 0.91 & \multicolumn{1}{c|}{40.4} & - & - & - & - & - & - \\
 & Low  & 0.37 & 0.93 & \multicolumn{1}{c|}{42.8}& 0.063 & 0.91 & \multicolumn{1}{c|}{40.0} & - & - & - & - & - & - \\
 & Medium & 0.76 & 0.93 & \multicolumn{1}{c|}{41.1} & 0.066 & 0.91 & \multicolumn{1}{c|}{39.6} & - & - & - & - & - & - \\
 & High & 0.84 & 0.86 & \multicolumn{1}{c|}{40.6} & 0.060 & 0.92 & \multicolumn{1}{c|}{40.9} & - & - & - & - & - & - \\
  \hline 
\multirow{4}{*}{\begin{tabular}[c]{@{}c@{}}MochiDiff\\ germline\\ (ours)\end{tabular}} &
  None  & NA & NA & \multicolumn{1}{c|}{NA} & 0.057  & 0.93 & \multicolumn{1}{c|}{61.3} & 0.19 & 0.27 & 0.22 & 0.96 & 36.1 & 24.3 \\
 & Low  & NA & NA & \multicolumn{1}{c|}{NA} & 0.031 & 0.91 & \multicolumn{1}{c|}{61.3} & 0.37 & - & - & - & 35.0 & 25.7 \\
 & Medium  & NA & NA & \multicolumn{1}{c|}{NA} & -0.007 & 0.86 & \multicolumn{1}{c|}{61.9} & 0.51 & - & - & - & 33.2 & 25.1 \\
 & High & NA & NA & \multicolumn{1}{c|}{NA} & -0.063 & 0.81 &  \multicolumn{1}{c|}{64.9} & 0.80 & 0.35 & 0.37 & 0.94 & 33.5 & 26.4 \\
 \hline 
\multirow{4}{*}{\begin{tabular}[c]{@{}c@{}}EvoProt\\ Grad \cite{emami2023plug} \end{tabular}} &
  None  & 0.10 & 0.87 & \multicolumn{1}{c|}{63.0} & 0.066 & 0.752 & \multicolumn{1}{c|}{67.3} & 0.21 & 0.26 & 0.23 & 0.97 & 31.2 & 21.7 \\
 & Low  & 0.21 & 0.83 & \multicolumn{1}{c|}{72.0} & -0.001 & 0.66 & \multicolumn{1}{c|}{85.3} & 0.23 & - & - & - & 30.5  & 22.1 \\
 & Medium  & 0.48 & 0.59 & \multicolumn{1}{c|}{84.8} & -0.029 & 0.55 & \multicolumn{1}{c|}{85.1} & 0.40 & - & - & - & 28.9 & 22.4 \\
 & High & 0.77 & 0.60 & \multicolumn{1}{c|}{85.2} & -0.037 & 0.51 & \multicolumn{1}{c|}{85.8} & 0.77 & 0.29 & 0.25 & 0.92 & 29.0 & 22.3 \\ 
 \hline 
    \multirow{1}{*}{\begin{tabular}[c]{@{}c@{}} OAS \end{tabular}} &
  NA  & 0.14 & 0.92 & \multicolumn{1}{c|}{60.5} & 0.068  & 0.92 & \multicolumn{1}{c|}{60.5} & 0.11 & 0.24 & 0.22 & 0.96 & 60.5 & NA \\
 \hline \\
\end{tabular}%
}
\vspace{-4pt}
  \caption{{\bf Benchmarking our diffusion model on generative tasks.}
    Comparison of methods across three conditional generation tasks: sampling antibodies of a specific V-gene class, sampling sequences with lower hydrophobicity, and designing improved binders to the antigen HGFR. For each task we report how well the classifier adheres to the condition, as well as the average similarity to the nearest-neighbor in OAS (NN, percent identity) as well as the diversity of the generated sequences (Div, pairwise Levenshtein distance). A set of randomly sampled sequences from OAS is also included to allow for comparison to the distribution of natural antibodies for each metric. 
    Germline diffusion cannot be applied to the V gene task because the V gene is already determined in the germline. 
    For the HGFR binding task, where a clinical antibody is used as a seed, distance to the seed is reported instead of nearest neighbor. Additionally, Boltz-2 multimer was used to generate additional binding (Boltz-2 p(bind), ipTM) and structure quality (pLDDT) metrics for orthogonal validation \cite{passaro2025boltz2}. Only a subset of methods were tested in this experiment and evaluated with Boltz-2 due to computational constraints. Samples size N=256 for V gene and hydrophobicity, N=128 for HGFR binding. 
    \vspace{-18pt}
    }
  \label{table:design}
\end{table}

Next, we evaluate non-germline prediction performance, where the most likely non-germline residue is predicted at each position the observed antibody sequence differs from the germline. 
As a proxy for downstream design tasks, this evaluates how accurately each model proposes biologically plausible sequence variants. 
For this task we also calculate a simple nearest-neighbor lower bound, where non-germline residues are predicted based on the observed sequence for the nearest neighbor in the training set. We also obtain an upper bound on performance by finding the nearest test-set neighbor by peeking at the ground-truth observed sequence, and using that to predict the non-germline residue. This represents the true biological variability in plausible non-germline residues at a given position. Finetuning on antibody specific data also improves performance on this non-germline residue task 
(Table \ref{table:bench}). 
All approaches significantly outperform the nearest neighbor baseline at 12.7\%, but fall well short of the 53.8\% upper bound on performance. Additionally, the degree of improvement over vanilla ESM-2 is less than the  perplexity improvements would suggest. This is in line with prior work that has demonstrated a dramatic germline bias in antibody pLMs, with the majority of improved perplexity obtained by antibody specific training deriving from simply memorizing the germline at the expense of performance on non-germline residues \cite{olsen2024ablang2, matsen2025separating}. Our germline diffusion model mitigates this bias by starting from the germline sequence and only capturing variations thereafter. This enables a dramatic improvement to 46.3\% accuracy on this non-germline residue prediction task. 

\begin{figure*}[!htbp]
  \centering
    \includegraphics[height=2.71in]{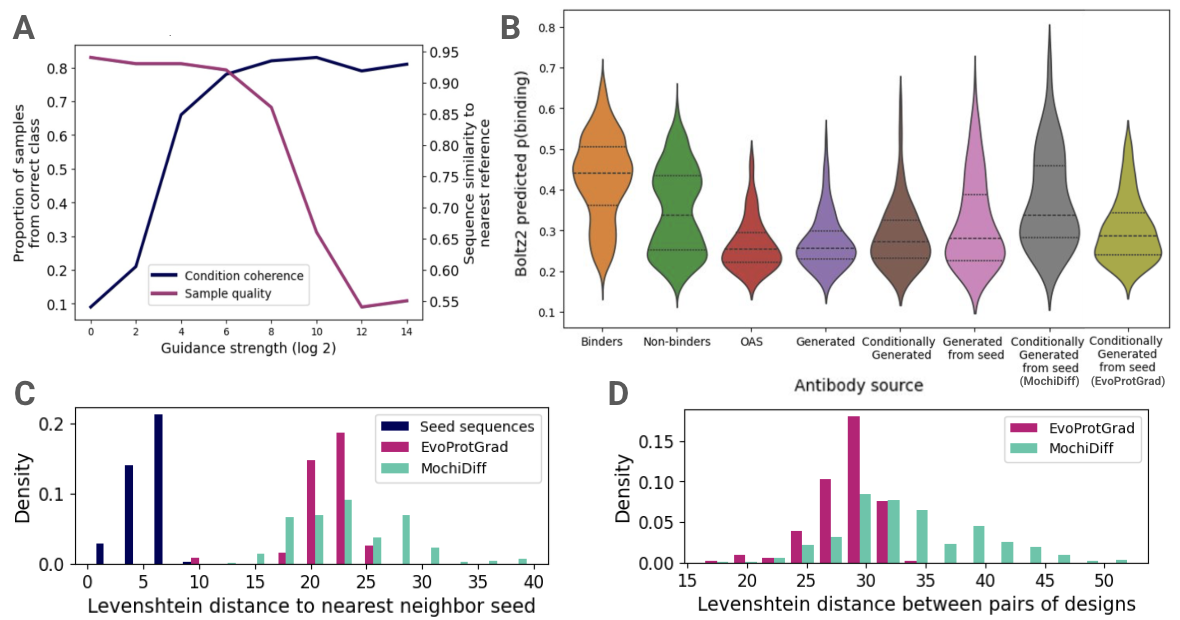}
  \caption{{\bf Conditional generation with germline diffusion}
    (A) Tradeoff between class coherence and sample quality when choosing a guidance strength for conditional generation. Antibody sequences conditionally sampled from one of the seven primary V-gene families using a pretrained classifier for guidance. (B)
    Orthogonal validation using Boltz-2 to predict structure-based binding affinity of sequences from experimentally derived binders (Orange), Non-binders (Green), random OAS antibodies (Red), unconditional (Purple) and conditional (Brown) germline diffusion, unconditional (Pink) and conditional (Gray) germline diffusion from seed, and conditional generation from seed with EvoProtGrad (Yellow). (C\&D) Histogram of Levenshtein distances between each design and its nearest neighbor (C) in the seed dataset of starting designs and (D) among other predicted designs. \vspace{-12pt} 
  }
  \label{fig:binders}
\end{figure*}

\paragraph{Antibody generation can be conditioned on arbitrary classifiers}
As a simple test of MochiDiff's ability to generate classifier-guided conditional samples, we consider the task of sampling a heavy chain sequence from one of the 7 main V gene families. Human heavy-chain variable region genes are classified into seven main classes based on sequence homology, which can easily be predicted from sequence with 100\% accuracy by a simple linear classifier trained on top of ESM-2 8M embeddings. 
Using this classifier to guide the diffusion sampling process, we observe a direct tradeoff between the strength of the guidance - and thus the conformity to the desired class - and the quality of the produced samples (Figure \ref{fig:binders}A).
When the classifier guidance is too low, samples are high quality, as approximated by their sequence similarity to their nearest reference in the train set. 
However, they do not conform to the desired class better than random chance. 
At the other extreme, when the classifier guidance is the highest, samples conform to the desired class 82\% of the time, but sampling ends up significantly off manifold. 
Manual tuning is required to find an appropriate tradeoff where sample quality and conditional adherence are both high. 

We next evaluate the more practically useful setting of sampling antibodies conditioned on hydrophobicity. Low surface hydrophobicity improves the developability and pharmacokinetics of monoclonal antibodies \cite{raybould2019five, jain2017biophysical}, making it a common target in directed evolution campaigns. For guidance, we use the same linear probe of ESM-2 as above, but trained on a developability dataset with in-silico hydrophobicity for $\sim$2 million antibodies \cite{bashour2024biophysical}. This hydrophobicity guidance model achieves near-perfect performance, with a test set Spearman correlation of 0.998. 
Reassuringly, for unconditionally generated sequences from our germline diffusion model the distribution of hydrophobicity almost perfectly match those for true sequences drawn from OAS (Figure \ref{fig:sampling}B). 
Applying classifier guidance markedly skews the distribution of predicted hydrophobicity lower while maintaining sample diversity comparable to the true OAS data distribution, demonstrating that we can successfully skew generation from our model towards properties of interest (Table \ref{table:design}). 
Comparing germline diffusion to the SEDD uniform model, the germline model shows much better adherence to the classifier guidance. We hypothesize that this is due to the germline model starting from a reference antibody sequence which is in-distribution for the pretrained classifier, unlike the uniform diffusion model, which starts from a fully noised sequence. As the guidance strength for the germline model is increased, we again see a tradeoff between sample quality and class adherence, with higher guidance strength decreasing hydrophobicity but increasing distance to training set sequences. However, when compared to EvoProtGrad on this same task, our model is Pareto dominant in this tradeoff, yielding samples which are both more plausible antibody sequences and have lower predicted hydrophobicity (Table \ref{table:design}). 

\vspace{-10pt}

\paragraph{Conditional generation of better HGFR binders}
Finally, we turn to a simulated \emph{in silico} directed evolution campaign to improve binding affinity against HGFR, a cell-surface receptor linked to cell growth and cancer, starting from the clinical antibody Emibetuzumab as a seed. To train a classifier for this task, we use a binding affinity dataset which contains 4000 random CDR-H3 variants of Emibetuzumab as well as experimentally determined binding labels for each sequence \cite{makowski2022cooptimization}, achieving an AUROC of 0.97 (Figure \ref{fig:guidclf}C).   
Generating conditional samples from our germline diffusion model guided by this classifier, we obtain new sequences with improved predicted binding affinity (Table \ref{table:design}). However, using the same classifier for both guidance and validation may provide an unrealistic view of performance - it is possible that we are generating samples which are simply adversarial against the classifier as opposed to having the desired property. 

To obtain orthogonal validation, we turn to an alternative structure-based binding affinity prediction using Boltz-2 \cite{passaro2025boltz2}. 
Evaluating Boltz-2's structural interface metrics we find that, while far from perfect, they are reasonably strong predictors of binding in the HGFR binding dataset (Figure \ref{fig:boltz2}, \ref{fig:binders}B). 
Looking at the predicted binding affinities for both random sequences in the OAS training set and unconditionally generated antibodies from our germline diffusion model, we see that both show a similar distribution of low predicted binding affinity (Figure \ref{fig:binders}). 
When we condition generation with our binding classifier, starting from OAS germlines, we see that the distribution of Boltz-2 binding scores shifts up somewhat. 
However, they are still much worse than the binders in the Emibetuzumab variant dataset. 
This is unsurprising, since most germlines used as a starting point here are far from strong HGFR binders. 
Biologically, only the best binders among the naive B-cell repertoire undergo affinity maturation to become strong binders. 
To address this, we instead use the 4000 Emibetuzumab variant sequences instead of random germline sequences as the starting point for the germline diffusion model. 
As expected, sampling unconditionally in this setup shifts binding affinity back down towards that observed for random OAS sequences. 
When we sample conditioned on binding, the Boltz-2 predicted binding affinity remains much higher on average, and two sequences receive the highest scores out of all sequences considered (Figure \ref{fig:binders}B). Additionally, these sequences still have high confidence Boltz-2 predicted structures (pLDDT, Table \ref{table:design}) indicating high-quality designs. 
To compare MochiDiff to an existing approach for classifier guided directed evolution, we test EvoProtGrad on the same task \cite{emami2023plug}, 
with AbLang-2 as our proposal pLM and our binding classifier as an additional expert. 
While sampling from seed sequences conditioned on HGFR binding affinity with EvoProtGrad does lead to somewhat higher Boltz-2 binding scores than OAS sequences, it significantly underperforms samples from our method (Figure \ref{fig:binders}B). 

Finally, we analyze the diversity in the designed sequences from each method in terms of Levenshtein distance both between pairs of designs (Figure \ref{fig:binders}D) and to the original seed (Figure \ref{fig:binders}C). 
We see that both EvoProtGrad and our diffusion model generate a highly diverse set of designs which all differ substantially from the starting seed. 
This indicates that our method successfully generates new binders which are on par with the original seed, while introducing significant additional sequence diversity. For directed evolution campaigns, where multiple metrics such as binding affinity, specificity, and developability properties are optimized jointly, this increased sequence diversity is highly desirable, allowing for new Pareto optimal points on the tradeoff between objectives to be explored.
\vspace{-8pt}

\section{Conclusion and future work}
\vspace{-2pt}
In this work, we present a new antibody specific pLM trained with discrete diffusion. We show that diffusion has favorable inductive biases for protein sequences and allows for flexible conditional generation with classifier guidance. We also present a novel modification to the diffusion process tailoring it specifically to antibody sequences, where the noise process has the germline sequence as an absorbing state. This setup dramatically mitigates the germline bias present in many antibody pLMs, improving its utility for downstream design tasks. Additionally, because the diffusion process is constrained to the space of plausible antibody sequences, it keeps guidance classifiers in-distribution. Based on our findings, other antibody pLMs may benefit from exploring alternative ways of conditioning generation on the germline, such as concatenating it to the input during masked language modeling. Furthermore, our framework has promising applications to other areas of biology, where evolutionary data would allow for training generic DNA or protein diffusion language models, as well as to other more disparate domains in physics and chemistry, where conditioning the diffusion process on specific initial conditions may be desirable. 

There remain a number of promising future directions building off of this work. First, it will be interesting to explore conditioning generation on multiple properties of interest at once, and evaluating whether MochiDiff is able to successfully balance potentially conflicting design objectives. To this end, another important step will be developing strategies to reduce the computational cost of classifier guidance. Currently guidance requires a number of classifier forward passes that is linear in both the sequence length and number of diffusion steps. This is tractable for the relatively small guidance classifiers considered here, but scaling up will require new optimizations to accelerate or approximate the current computation. Finally, while we perform multiple forms of \emph{in silico} evaluation of our model, future work applying it to real design campaigns with wet-lab validation will provide a more comprehensive view of MochiDiff's utility in practice. 

\bibliographystyle{plain}
\bibliography{refs}
\clearpage

\appendix

\setcounter{figure}{0}
\setcounter{table}{0}
\setcounter{section}{0}
\renewcommand{\thefigure}{S\arabic{figure}}
\renewcommand{\thesection}{S\arabic{section}}
\renewcommand{\thetable}{S\arabic{table}}

\section{Supplementary Figures}

\begin{figure*}[ht!]
  \centering
    \raisebox{118pt}{\makebox[0pt][l]{\textbf{(A)}}}%
    \includegraphics[height=1.56in]{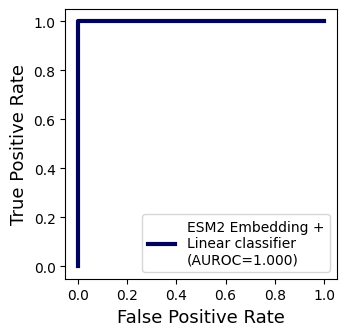} 
    \raisebox{118pt}{\makebox[0pt][r]{\textbf{(B)}}}%
    \includegraphics[height=1.56in]{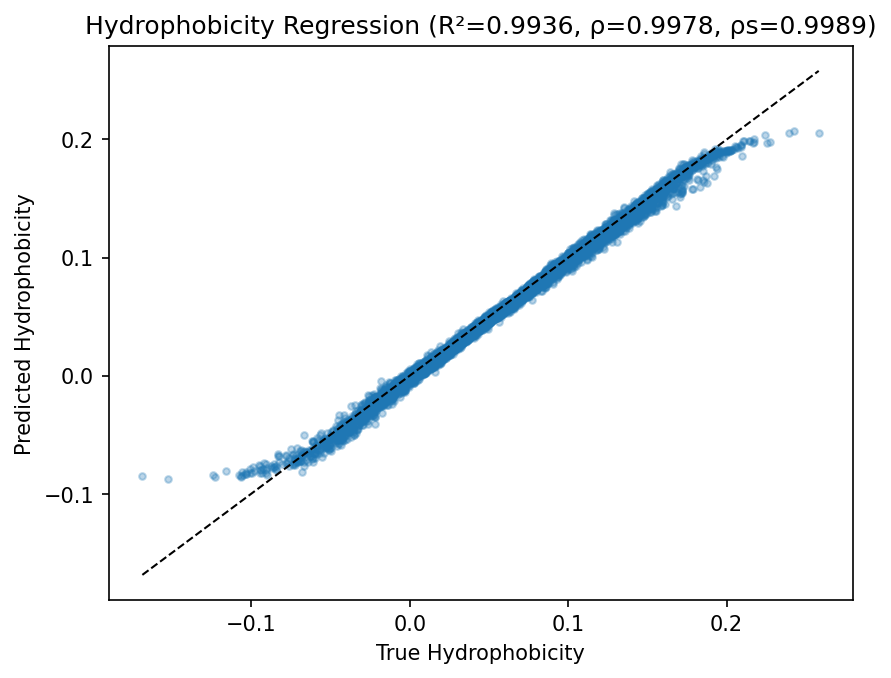} 
    \raisebox{118pt}{\makebox[0pt][r]{\textbf{(C)}}}%
    \includegraphics[height=1.56in]{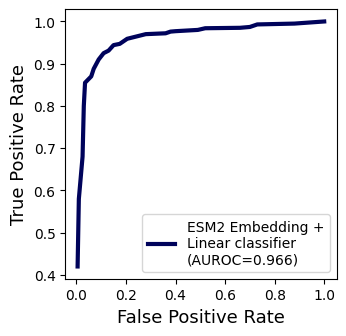} 
  \caption{{\bf Performance of guidance models} Performance of the ESM-2 linear probe models used for classifier guided conditional generation for (A) predicting the V gene class (B)  predicting hydrophobicity (C) separating strong versus weak binders to HGFR.
  }
  \label{fig:guidclf}
\end{figure*}

\begin{figure*}[ht!]
  \centering
    \raisebox{135pt}{\makebox[0pt][l]{\textbf{(A)}}}%
    \includegraphics[height=1.82in]{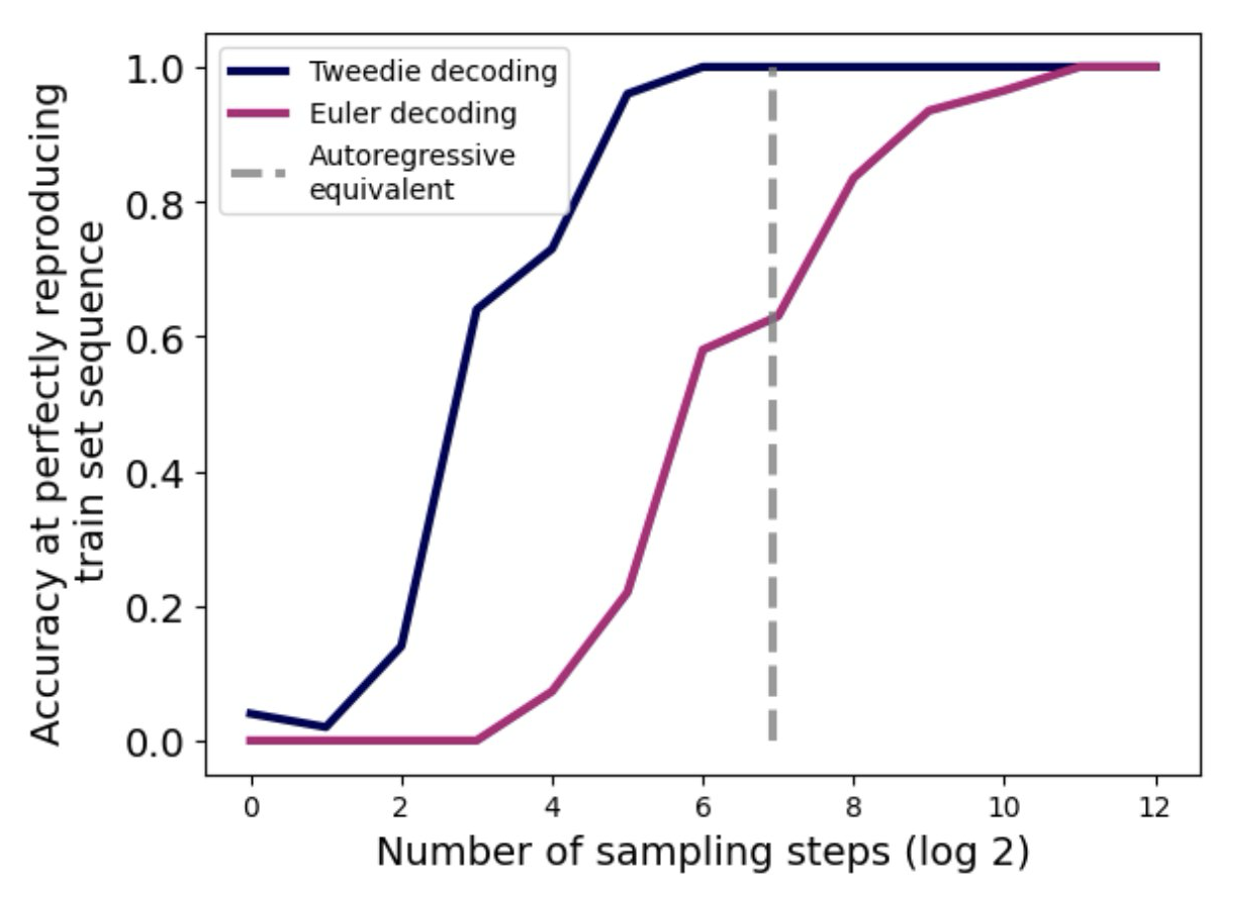}  
    \raisebox{135pt}{\makebox[0pt][r]{\textbf{(B)}}}%
    \includegraphics[height=1.9in]{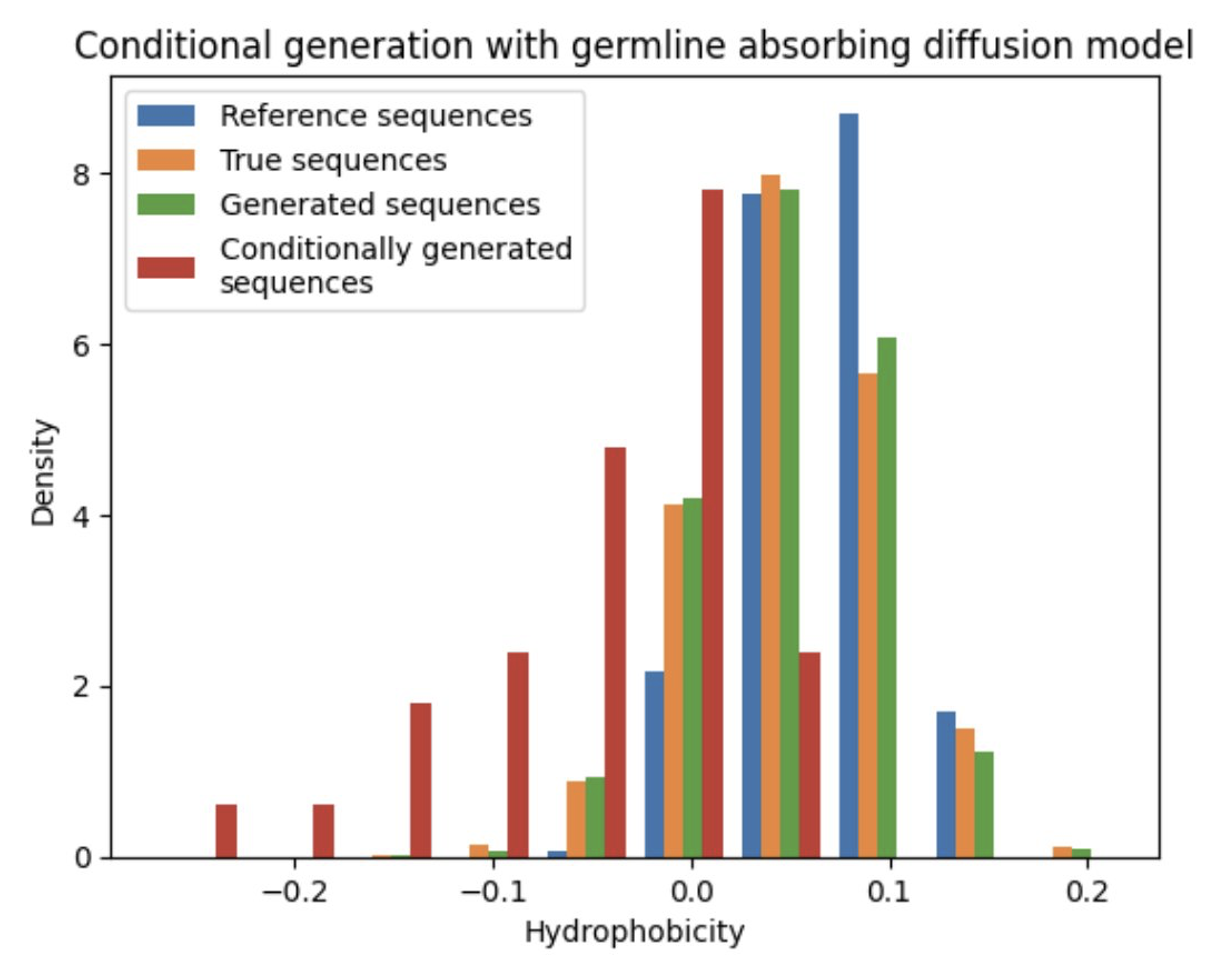} 
  \caption{{\bf Sampling from antibody diffusion model}
    (A) Tradeoff between number of diffusion sampling steps and sample quality for the two diffusion strategies, as measured by ability of an overfit model to reproduce test-set sequences. The number of forward passes required by an autoregressive model for the same length sequence is indicated with a dashed gray line. (B) Conditional generation guided by a hydrophobicity property prediction model can skew generation towards antibodys with good developability properties.
  }
  \label{fig:sampling}
\end{figure*}

\begin{figure*}[htb!]
  \centering
    \includegraphics[height=3in]{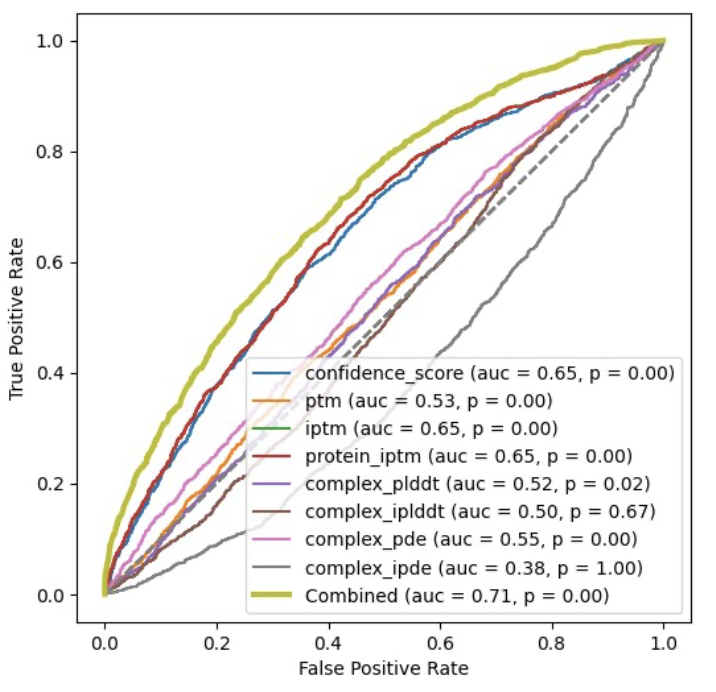}  
  \caption{{\bf Boltz-2 multimer binding prediction}
    ROC curves evaluating how well each structure-based metric calculated by Boltz-2 is able to separate strong binders vs. weak binders on the Makowski et al. HGFR dataset. Cross-validated performance of a combined classifier trained on all metrics is shown in yellow.  
  }
  \label{fig:boltz2}
\end{figure*}

\begin{figure*}[htb!]
  \centering
    \includegraphics[height=2.1in]{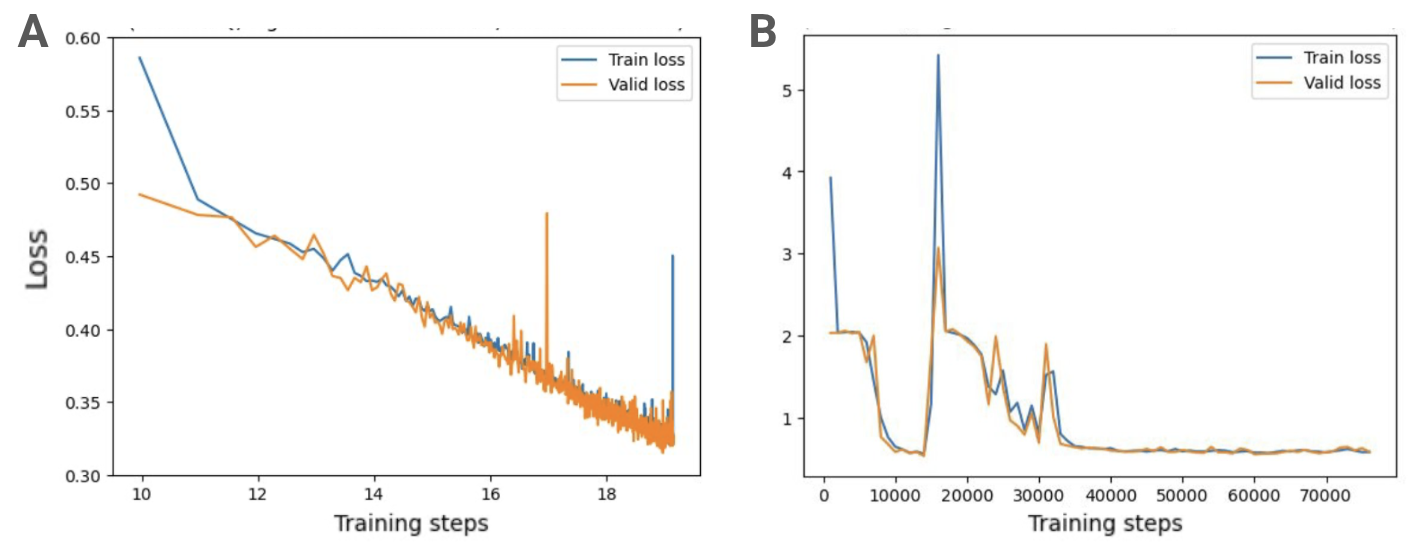}  
  \caption{{\bf ESM-2 initialization improves training}
    Training curves for our absorbing diffusion model initialized (A) with ESM-2 pretrained weights and (B) random weights. ESM-2 initialization dramatically improves training and performance.   
  }
  \label{fig:training_curve}
\end{figure*}

\clearpage 

\section{Additional Explanations and Experimental Details}

\subsection{Training data preparation}
The Observed Antibody Space dataset (OAS), which contains $>1.1$ billion sequenced B-cell receptors, is the \textit{de facto} standard for training antibody protein language models. 
In order to retain germline sequence and and V gene class annotations for each sequence, we needed to re-process the data and create new train-test splits. 
Thus, we downloaded each of the experimental raw files from the OAS data portal 
(\href{https://opig.stats.ox.ac.uk/webapps/oas/oas_unpaired/}{https://opig.stats.ox.ac.uk/webapps/oas/oas\_unpaired/}) 
filtered for unpaired human heavy chain sequences from Peripheral Blood Mononuclear Cells (PBMC). 
This yields $\sim$1.1 billion unfiltered sequences. These were then filtered using the same criteria as Olsen et al. 2023, namely by requiring complete V(D)J annotations, no early stop codon or frameshift, productive sequences (assigned by IgBlast), no unassigned `X` residues in the sequence, and no missing conserved Cysteines. 
After applying these filtering rules there are 337 million total sequences remaining. 
To yield the final dataset of sequences, we then perform clustering on the filtered data using mmseqs linclust \cite{steinegger2017mmseqs2, steinegger2018linclust}. 
To ensure high sequence diversity, we apply a strict 70\% sequence identity cutoff. We then filter the cluster sequences based on a 80\% germline sequence identity threshold, ensuring that only distinct B-cell lineages are included in the data. 
This gives a final dataset of $\sim$25.6M sequences. 

\subsection{Model training details}
All of the models trained are based on the ESM-2 650M architecture published on Hugging Face \cite{lin2023evolutionary}. For our antibody-finetuned MLM baseline in Table \ref{table:bench}, we finetune the pretrained model with a batch size of 1024, learning rate of 2e-4, and the same masking setup used during pretraining (15\% mask rate, 80\% mask token, 10\% randomized, 10\% unchanged). Training was stopped early when validation loss diverged after $\sim$117,000 training steps, corresponding to $\sim$4.8 epochs. 

For our three diffusion models, we trained with a batch size of 256, learning rate of 2e-4, gradient norm clipped to 2, an exponential moving average of 0.9999, and a linear warmup schedule. Models were trained for 524,288 training steps, corresponding to roughly 5.4 epochs. Model training exhibits a typical log-linear relationship between training examples and loss, and showed no signs of overfitting (Figure \ref{fig:training_curve}A) Training examples were sampled from uniform time points on [0,1] and noised with a log-linear noise schedule $\sigma(t) = \ln(1-(1-\epsilon)t)$, where $\epsilon = 10^{-3}$ was included for numerical stability around $t=1$. To simplify training our absorbing diffusion models, we use the reparameterized absorbing discrete diffusion (RADD) trick presented by Ou et al \cite{ou2024absorbing} to give our model only analytical time dependence, removing the need to parameterize our network with the timestep $t$. For the uniform diffusion model, we use in-context conditioning with a sinusoidal time embedding with learned frequencies. This requires fewer architecture changes, allowing pretrained ESM2 to more easily be used out of the box, but future work may benefit from the AdaLN approach widely used in diffusion transformers \cite{peebles2023dit}.

Due to resource limitations, we did not perform a hyperparameter or architecture search for any of our models. Model architecture was chosen to match ESM-2 for convenience, and hyperparameters were set to common defaults based on the literature. 

To evaluate the impact of initializing the model using pre-trained ESM-2 weights, we tested the training our [MASK] absorbing diffusion model from scratch. We observe that training is much more unstable, and eventually gets stuck in a local minima at a relatively high loss of $\sim$0.46 (Figure \ref{fig:training_curve}B). Additionally, the initial loss at the start of training is much higher. This indicates that pre-trained generic protein representations from ESM-2 are highly beneficial for this task, even though the model is pre-trained using a different strategy.  

All model training was performed on AWS EC2 p5.48xlarge instances with 8 A100 80GB GPUs and 2 TiB of instance memory. 
DDP with gradient accumulation was used for distributed training to allow for batch sizes larger than would fit in a single GPUs memory.

\subsection{Language modeling evaluation}
To benchmark our models performance against existing pLMs, we compare to ESM-2 \cite{lin2023evolutionary}, AbLang-2 \cite{olsen2024ablang2}, and IgLM \cite{shuai2023generative}, each used under appropriately their non-commercial licenses (MIT, BSD-3, JHU respectively). Pretrained published weights for each model were used. Due to differing train/test splits used to train each model, there is some concern over data leakage in our evaluation, which may benefit some models more than others. 

\paragraph{Perplexity calculation} To calculate perplexity for masked language models, we adopt the pseduo-perplexity approach, where each token in a test set sequence $x$ of length $N$ is iteratively masked and predicted by the model:
$$\text{Perplexity}(x) = \exp\left( -\frac{1}{N} \sum_{i=1}^{N} \log P_\theta(x_i \mid x_{\setminus i}) \right)$$
where $x_{\setminus i}$ is the sequence $x$ with the token at position $i$ masked. 

For diffusion models, we calculate an upper bound on test set perplexity using the same ELBO on negative log likelihood used during training: 
\begin{multline*}
    \text{Perplexity}(x)
    \;\leq\; \exp \biggl(
    \int_0^T
        \mathbb{E}_{x_t \sim p_{t|0}(\cdot\mid x)}
        \sum_{y \neq x_t}
        Q_t(x_t, y)
        \biggl[
            s_{\theta}(x_t, t)_y - \\
            \frac{p_{t|0}(y \mid x)}{p_{t|0}(x_t \mid x)}
              \log s_{\theta}(x_t, t)_y
            + K\!\left(\frac{p_{t|0}(y \mid x)}{p_{t|0}(x_t \mid x)}\right) \biggr] dt
    \;+\;
    D_{\mathrm{KL}}\!\bigl(p_{T|0}(\cdot\mid x)\;\Big\|\;p_{\mathrm{base}}\bigr) \biggr)
\end{multline*}
 
where $K(a) = a(\log a - 1)$ is a normalizing constant function and the KL divergence term captures the distance between fully noised sequence and the base distribution (Either a uniform distribution or a single absorbing state), which is negligible in practice due to the design of the noise process.

\paragraph{Non-germline accuracy calculation} Non-germline accuracy measures how often a model is able to predict the observed residue at positions where the observed sequence differs from the germline. This is an important metric which captures how well a model has learned to capture the processes of peripheral/central tolerance and affinity maturation during B-cell development. To calculate this accuracy for the masked language models, we mask each non-germline residue in each test set sequence individually, and ask the model to predict the most likely non-germline residue at the masked position (the logit corresponding to the germline residue, which is usually the most likely, is excluded). To predict non-germline residues with our diffusion models, we use a similar setup: for each non-germline position we provide as input a sequence with just that position noised (Uniform, [MASK]'d, or replaced with the germline) and use the concrete score predicted by the model's logits to assign the most likely non-germline residue (again excluding the germline amino acid). For the uniform diffusion model, which is parameterized with the timestep, we use $t = \frac{1}{128}$, indicating this is the final denoising step. 

\subsection{Sampling from trained models}
To sample from out trained diffusion models, we experimented with both the Euler and Tweedie decoding strategies described by Lou et al \cite{lou2024discrete}. As an initial experiment to compare the two strategies and evaluate the tradeoff between the number of diffusion sampling steps and the quality of the resulting generation, we overfit our SEDD uniform model to a small training set of 1,000 sequences, and evaluate how well the model can reproduce a train set sequence. 
As expected, we find that increasing the number of diffusion sampling steps monotonically increases sample quality. 
We also find that Tweedie decoding is about 8-32x more efficient than the Euler method, generating perfect reproductions of train sequences at 64 sampling steps for Tweedie decoding and 2048 sampling steps for Euler decoding (Figure \ref{fig:sampling}A). This result is in line with the findings reported by Lou et al. The result also demonstrates that sampling from our diffusion model with Tweedie decoding is approximately as efficient as autoregressive alternatives, which require one forward pass per residue, or $\sim$120 for heavy chain variable domains. Based on these findings, we use Tweedie decoding with 128 sampling steps for all generative experiments. 

\subsection{Downstream experiments}
We test our models on three downstream conditional generation tasks. 
For each task, the guidance model consisted of fine-tuning the ESM-2 protein language model (150M parameter variant, esm2\_t30\_150M\_UR50D), with a single linear output head appended to the pretrained transformer backbone. 
The models were trained with a batch size of 128, a learning rate of 1e-3, and early stopping (patience of 3 evaluations, threshold 1e-4) applied on validation loss evaluated every 50 steps. The best checkpoint by validation loss was retained. For the two classification tasks we use a binary cross-entropy loss and asses model performance on the held out test set based on accuracy and AUROC. For the regression task, we use MSE loss and calculate $R^2$, Pearson correlation, and Spearman rank correlation. 

During conditional generation, there is an inherent tradeoff between sample quality and class coherence that can be tuned by adjusting the guidance strength. For each task, we test three different guidance strengths, corresponding to low, medium, and high guidance, in order to demonstrate a range of values on the frontier of this tradeoff . 
Empirically, we found that different tasks (and different guidance models) require different guidance strengths for reasonable results. Thus, we use [8, 64, 512] as guidance strength for the V gene experiment, [32, 64, 128] for the hydrophobicity experiment, and [1, 4, 16] for the HGFR binding experiment. For each experiment, we also perform unconditional sampling (guidance strength 0). Properties from unconditional samples should closely match those for random OAS sequences. 

\paragraph{EvoProtGrad}
To perform conditional generation with EvoProtGrad on these same tasks, we use Ablang-2 mutant marginal scoring strategy as the pLM expert for proposing mutations. Guidance classifiers are provided as additional experts implemented by extending the provided generic onehot downstream regression/classifier classes. Sampling was performed with 5 parallel chains, 200 chains, and 20 max mutations, with the best scoring sample across all chains from each run kept. Max mutations of 20 was selected to obtain sequences with approximately the same diversity as was observed from our germline diffusion model (Figure \ref{fig:binders}C\&D). 

Similar to our diffusion models, the strength of classifier guidance for EvoProtGrad needs to be tuned by manually adjusting the temperature for each expert. For our V gene class experiments, we used [4, 16, 64], for hydrophobicity we used [16, 32, 64], and for HGFR binding we used [2, 8, 32]. These values are not directly comparable accross models or tasks. 

\subsubsection{V gene class experiment}
Human heavy-chain variable region genes (\(V_{H}\)) are classified into seven main clans/families based on sequence homology. We generate balanced training data for the task of predicting V gene class from sequence by randomly sampling from  OAS to generate a train/vaidation/test sets with 70000/7000/7000 sequences with an equal number from each class. As may be expected, this task proves very easy, with the classifier on top of ESM2 embeddings achieving perfect accuracy (Figure \ref{fig:guidclf}A). 

\paragraph{Hydrophobicity experiment}
Hydrophobicity refers to the thermodynamic tendency of nonpolar amino acid residues to bury within the protein core thus conferring structural stability through the hydrophobic effect. In antibody design, managing surface hydrophobicity is critical, as excessive hydrophobic patches on the variable domain surface are associated with poor developability outcomes. To train our hydrophobicity model for guidance, we use a dataset from Bashour et al. with hydrophobicity values (transfer free energy measured in kcal/mol) for $\sim$2 million antibodies \cite{bashour2024biophysical}. We downsample this data randomly to 50000/10000/10000 train/validation/test set sequences. The model achieved near-perfect performance on the test set, with R2 = 0.99, Pearson $\rho$ = 0.99, and Spearman $\rho$ = 0.99 (\ref{fig:guidclf}B).


\paragraph{HGFR experiment}
Our HGFR binding classifier is trianed on a Makowski et al dataset of 4000 random CDR-H3 variants of Emibetuzumab, with experimentally determined binary high/low binding affinity labels for each sequence \cite{makowski2022cooptimization}. We split this data into 3200/400/400 sequence train/valid/test splits, and obtain an AUROC of 0.97 for a classifier that seperates good from weak binders (Figure \ref{fig:guidclf}).   

\paragraph{Boltz-2}
As orthogonal validation of our generated HGFR binder designs, we use multimer predictions. Boltz-2 is a structure-based foundation model which jointly models complex structures and binding affinities for protein complexes. We generate one multimer structure for each design, providing the HGFR sequence as a second entity and providing it's PDB structure as a template. Prediction is run with use\_msa\_server and use\_potentials set to true, and all other parameters set to defaults. The resulting Boltz-2 predictions include 8 metrics which capture the quality/strength of the multimer structure. Evaluating these metrics on the test set sequences, we find that ipTM, complex\_ipde, and confidence\_score are all weakly predictive of experimental binding labels (Figure \ref{fig:boltz2}). Training a logistic regression model on these 8 metrics further increases performance somewhat to an AUROC of 0.71. Along with ipTM, we use the likelihood from this logistic regression model as our p(binding) for Boltz-2 in Table 2 and Figure 2.

\end{document}